\documentclass{article}

\usepackage[preprint]{neurips_2024}

\usepackage[utf8]{inputenc} 
\usepackage[T1]{fontenc}    
\usepackage{hyperref}       
\usepackage{url}            
\usepackage{booktabs}       
\usepackage{amsfonts}       
\usepackage{nicefrac}       
\usepackage{microtype}      
\usepackage{xcolor}         
\usepackage{tikz}
\usepackage[linesnumbered,ruled]{algorithm2e}
\usepackage{subcaption}
\usepackage{tikz}
\usetikzlibrary{shapes.geometric, arrows, positioning, fit, calc, trees, decorations.pathreplacing}
\usepackage{graphicx}

\title{Recent advancements in LLM Red-Teaming: \\Techniques, Defenses, and Ethical Considerations}

\author{%
  Tarun Raheja\\
  Independent Researcher\\
  San Francisco, USA\\
  \texttt{traheja@alumni.upenn.edu} \\
  \And
  Nilay Pochhi\\
  Independent Researcher\\
  San Francisco, USA\\
  \texttt{pochhi.nilay@gmail.com} \\
  \And
  F.D.C.M. Curie\\
  Independent Research Associate\\
  San Francisco, USA\\
  \texttt{} \\
}

\begin{document}

\maketitle

\begin{abstract}
Large Language Models (LLMs) have demonstrated remarkable capabilities in natural language processing tasks, but their vulnerability to jailbreak attacks poses significant security risks. This survey paper presents a comprehensive analysis of recent advancements in attack strategies and defense mechanisms within the field of Large Language Model (LLM) red-teaming. We analyze various attack methods, including gradient-based optimization, reinforcement learning, and prompt engineering approaches. We discuss the implications of these attacks on LLM safety and the need for improved defense mechanisms. This work aims to provide a thorough understanding of the current landscape of red-teaming attacks and defenses on LLMs, enabling the development of more secure and reliable language models.
\end{abstract}

\section{Introduction}

Large Language Models (LLMs) have shown immense potential in various natural language processing tasks. However, the safety and security of LLMs remain a major concern, as these models can be exploited to generate harmful, unethical, or biased content. This has led to an active research area focused on red-teaming LLMs, which involves probing and evaluating their vulnerabilities to adversarial attacks. 

\section{Automated Red-Teaming for LLMs}

Red-teaming is a crucial step for both model alignment and evaluation, but manual red-teaming is labor-intensive and difficult to scale. This has led to the development of automated red-teaming techniques, which automatically generate adversarial prompts or inputs to elicit undesirable responses from LLMs. \citet{red_perez} was one of the first to explore automated red-teaming, demonstrating that another LLM could be used to generate test cases that uncover harmful outputs from a target LLM.

\subsection{Reinforcement Learning based Red-Teaming}

Many automated red-teaming methods rely on reinforcement learning (RL). In these methods, an attacker language model is trained to generate prompts that maximize the likelihood of eliciting undesirable responses from the target LLM, often measured by an auxiliary safety classifier. 

\cite{learning_lee} proposes using GFlowNet fine-tuning followed by a smoothing phase to train the attacker model, aiming to generate diverse and effective attack prompts. This method addresses the mode collapse issue observed in previous RL-based approaches. Similarly, \cite{curiosity_hong} leverages curiosity-driven exploration to enhance the coverage and diversity of generated test cases, leading to a more comprehensive assessment of the target LLM's vulnerabilities.  \cite{diver_zhao} further introduces DiveR-CT, which relaxes conventional constraints on objective and semantic rewards to enhance the diversity of generated adversarial prompts. \cite{dart_jiang} proposes DART, which uses a deep adversarial interaction between a red team LLM and a target LLM to discover and address safety vulnerabilities in an iterative manner.

\subsection{Black-box Red Teaming}

Several studies focus on black-box red-teaming, where the attacker has limited access to the target LLM. This is particularly important as many real-world LLM APIs are black-box. 

\cite{query_lee} utilizes Bayesian optimization to efficiently discover diverse failure cases with a limited query budget. \cite{ferret_pala} improves upon Rainbow Teaming by introducing a scoring function to rank and select the most effective adversarial prompts, achieving higher attack success rates and efficiency. \cite{redagent_xu} presents RedAgent, a multi-agent LLM system that models jailbreak strategies and leverages them to generate context-aware jailbreak prompts for targeted attacks on LLM applications.

\subsection{Prompt Engineering and Optimization}

Prompt engineering plays a crucial role in designing effective jailbreak prompts. 

\cite{attack_deng} combines manual and automatic methods to generate high-quality attack prompts, leveraging LLMs to mimic human-generated prompts through in-context learning.  \cite{advprompter_paulus} introduces a novel algorithm that utilizes another LLM, called AdvPrompter, to generate human-readable adversarial prompts by optimizing for effectiveness and speed. \cite{maatphor_salem} proposes Maatphor, a tool that assists defenders in performing automated variant analysis of prompt injection attacks. It generates variants of existing attack prompts to test defenses against a wider range of potential attacks. \cite{goal_liu} presents a framework for constructing goal-oriented prompt attacks, aiming to induce LLMs to generate unexpected outputs by leveraging carefully crafted prompt templates. 

\subsection{Transferability and Generalization}

The transferability and generalization of jailbreak attacks across different LLMs are important considerations for red-teaming. 

\cite{a_ding} proposes ReNeLLM, a framework that leverages LLMs themselves to generate generalized jailbreak prompts, achieving higher attack success rates and efficiency. \cite{gradient_wichers} presents Gradient-Based Red Teaming (GBRT), an automatic method that utilizes gradients to generate diverse prompts that trigger unsafe responses, even when the target LLM has been safety-tuned. \cite{h4rm3l_doumbouya} introduces h4rm3l, a dynamic benchmark of composable jailbreak attacks that leverages a domain-specific language and program synthesis to generate novel and effective attack prompts.  \cite{easyjailbreak_zhou} presents EasyJailbreak, a unified framework that simplifies the construction and evaluation of various jailbreak attacks, enabling researchers to easily combine novel and existing components.  \cite{jailbreaker_deng} proposes Jailbreaker, an automated method that leverages time-based characteristics of LLM generation to identify and bypass defenses employed by mainstream chatbot services.

\section{Novel Attack Strategies and Benchmarking}

\begin{figure}
  \centering
  \includegraphics[width=\textwidth]{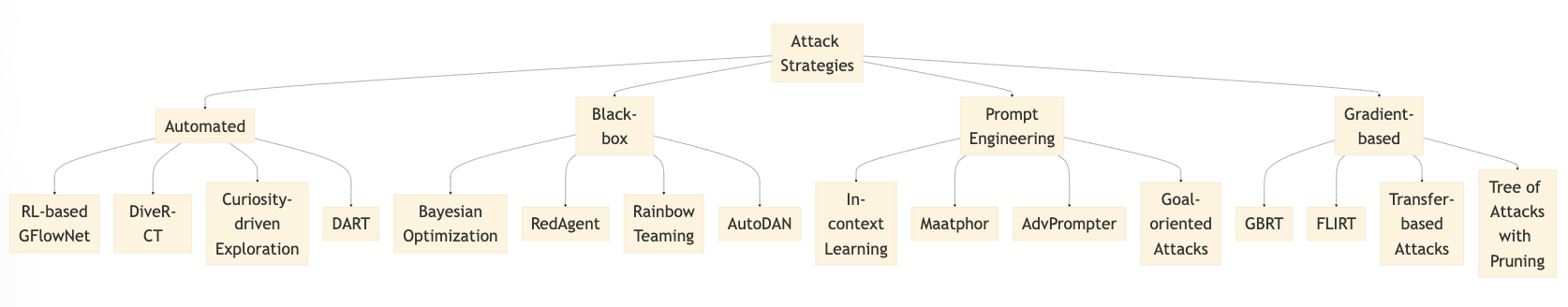}
  \caption{Attack Strategies}
  \label{fig:attack}
\end{figure}

\begin{figure}
  \centering
  \includegraphics[width=\textwidth]{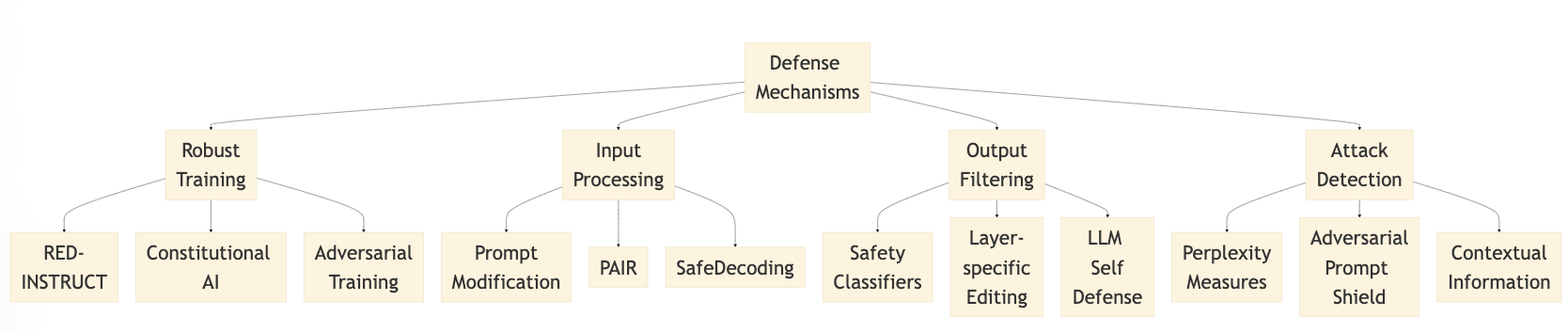}
  \caption{Defense Strategies}
  \label{fig:defense}
\end{figure}

Researchers continue to explore novel attack strategies to push the boundaries of LLM red-teaming. Recent advancements in LLM red-teaming have led to the development of innovative attack strategies, each exploiting unique vulnerabilities in language models:
Recent research has introduced several innovative approaches to challenge and evaluate the security of Large Language Models (LLMs). These novel attack strategies exploit various vulnerabilities and characteristics of LLMs, pushing the boundaries of red-teaming methodologies.

The Tastle Framework, proposed by \cite{tastle_xiao}, leverages LLMs' distractibility and over-confidence. By employing malicious content concealment and memory reframing techniques, Tastle successfully jailbreaks language models, exposing critical weaknesses in their defense mechanisms. Addressing the issue of unrealistic attack prompts, \cite{astprompter_hardy} introduced ASTPrompter. This method utilizes a novel reinforcement learning formulation to generate prompts that are both toxic and likely to occur in real-world scenarios, thereby increasing the practical relevance of red-teaming exercises. Exploiting the concept of social facilitation, the Social Prompt (SoP) method developed by \cite{sop_yang} generates and optimizes jailbreak prompts using multiple character personas. This approach taps into LLMs' susceptibility to social influences, revealing a unique vulnerability in their decision-making processes. \cite{gptfuzzer_yu} introduced GPTFuzz, an automated fuzzing framework that generates jailbreak templates by mutating human-written seed templates. This method exposes LLMs' sensitivity to subtle prompt variations, highlighting the need for more robust input processing mechanisms. The WordGame strategy, proposed by \cite{wordgame_zhang}, takes a different approach by simultaneously obfuscating queries and responses. This method effectively bypasses safety alignment measures in LLMs, demonstrating the potential vulnerabilities in current content filtering systems. Taking a ground-up approach, the Explore Framework developed by \cite{explore_casper} starts by exploring model behavior to establish definitions of undesirable outputs. It then leverages these insights to develop effective red-teaming strategies, providing a more comprehensive understanding of LLM vulnerabilities. Building upon previous work, \cite{masterkey_deng} introduced MASTERKEY, which extends the Jailbreaker framework by incorporating time-sensitive methods to circumvent LLM defenses. This automated approach demonstrates potential temporal vulnerabilities in chatbot security measures, opening new avenues for security research.

\subsection{Evaluation Frameworks and Benchmarks}

Standardized benchmarks and evaluation frameworks are crucial for comparing different red-teaming methods and tracking progress in LLM security. 

\cite{harmbench_mazeika} introduces HarmBench, a standardized evaluation framework for both automated red-teaming and robust refusal. This framework enables the systematic comparison of various attack and defense methods. \cite{evil_tian} proposes Evil Geniuses, a method for evaluating the safety of LLM-based agents by generating prompts that target vulnerabilities specific to different roles and interaction environments. 

\section{Defense Strategies and Mitigation Techniques}

Defending against jailbreak attacks and enhancing the robustness of LLMs are crucial aspects of responsible AI development. 

\cite{an_xu} introduces PromptAttack, a prompt-based adversarial attack that leverages the LLM itself to generate adversarial samples. \cite{prp_mangaokar} proposes PRP, a novel attack strategy that targets guard models, the additional layer of defense used in some LLMs, highlighting the need for more robust defenses. \cite{no_srivastava} builds upon previous work to develop a pipeline for automated test case generation, experimenting with different target LLMs and red classifiers to enhance the effectiveness of red-teaming. \cite{improved_jia} investigates techniques to improve the efficiency of optimization-based jailbreaking methods like GCG, proposing diverse target templates and adaptive optimization strategies. \cite{autodan_liu} introduces AutoDAN, an automated method that generates stealthy and semantically meaningful jailbreak prompts using a hierarchical genetic algorithm.

\subsection{Defenses based on Decoding, Prompt Modification and Alignment}

Several defense strategies focus on modifying the decoding process or altering prompt inputs to mitigate jailbreak attacks. 

\cite{enja_zhang} proposes EnJa, an ensemble jailbreak approach that combines prompt-level and token-level attacks for improved effectiveness. \cite{catastrophic_huang} investigates generation exploitation attacks that disrupt LLM alignment by manipulating decoding hyper-parameters and sampling methods, highlighting the need for more comprehensive alignment procedures. \cite{troublellm_xu} introduces TroubleLLM, an LLM designed for generating controllable test prompts for evaluating LLM safety issues, enabling more focused testing. \cite{kov_moss} utilizes Monte Carlo Tree Search to find harmful behaviors in black-box LLMs, emphasizing the importance of query efficiency in red-teaming. \cite{open_lapid} proposes a genetic algorithm-based approach for generating universal adversarial prompts that can jailbreak black-box LLMs. \cite{improved_li} leverages ideas from transfer-based attacks in image classification to enhance the effectiveness of adversarial prompt generation against LLMs. \cite{flirt_mehrabi} proposes FLIRT, a feedback loop in-context red-teaming method that utilizes in-context learning to automatically learn effective adversarial prompts for text-to-image models. \cite{tree_mehrotra} presents Tree of Attacks with Pruning (TAP), an automated method that utilizes an LLM to iteratively refine prompts using tree-of-thought reasoning, achieving efficient jailbreak with limited queries. \cite{red_poria} proposes RED-INSTRUCT, an approach for LLM safety alignment that leverages a chain of utterances to collect harmful questions and conversational data for safety-focused fine-tuning. \cite{rainbow_samvelyan} introduces Rainbow Teaming, a quality-diversity search-based approach for generating diverse and effective adversarial prompts, demonstrating its applicability across various domains. 

\subsection{Prompt Structure and Obfuscation}

Understanding and exploiting the structural characteristics of prompts are crucial for both attack and defense strategies. 

\cite{alpaca_kassem} investigates the use of instruction-based prompts to uncover LLM memorization, showing that these prompts can expose pre-training data more effectively than direct training data probing. \cite{exploiting_li} explores the impact of uncommon text-encoded structures (UTES) on jailbreak attacks, demonstrating that prompt structure significantly contributes to attack effectiveness. \cite{rl_chen} proposes RL-JACK, a reinforcement learning-powered black-box jailbreaking attack that formulates prompt generation as a search problem and optimizes it using a customized RL approach.  \cite{qroa_brunel} introduces QROA, a black-box attack that iteratively updates tokens to maximize a designed reward function, demonstrating the feasibility of black-box optimization for jailbreaking.

\subsection{Multimodal and Multilingual Jailbreaking}

As LLMs become increasingly multimodal and multilingual, new attack vectors and vulnerabilities emerge. 

\cite{jailbreaking_wu} investigates vulnerabilities in GPT-4V by exploiting system prompt leakage, demonstrating the potential risks of multimodal attacks. \cite{figure_lin} introduces Analyzing-based Jailbreak (ABJ), a method that exploits LLMs' analytical and reasoning capabilities to induce harmful behavior. \cite{adappa_lv} proposes AdaPPA, a jailbreak attack approach that utilizes the model's instruction-following and narrative-shifting capabilities to generate harmful content.  \cite{drattack_li} introduces DrAttack, a framework that decomposes and reconstructs prompts to obfuscate malicious intent, achieving higher attack success rates with fewer queries.  \cite{autobreach_chen} utilizes wordplay and LLM's summarization capabilities to generate diverse universal mapping rules for crafting adversarial prompts, improving the efficiency of jailbreaking. 

\subsection{False Refusals and Over-Refusals}

While safety alignment is crucial for preventing harmful outputs, it can also lead to over-refusal, where LLMs reject even harmless prompts. 

\cite{automatic_an} focuses on generating pseudo-harmful prompts to evaluate false refusals in LLMs, highlighting the trade-off between minimizing false refusals and improving safety. \cite{cognitive_xu} investigates cognitive overload attacks that target the cognitive structure and processes of LLMs, demonstrating their vulnerability to multilingual overload, veiled expression, and effect-to-cause reasoning. \cite{lockpicking_li} introduces JailMine, a token-level manipulation approach for jailbreaking that automates the process of eliciting malicious responses, addressing concerns about scalability and efficiency. \cite{autojailbreak_lu} presents AutoJailbreak, a framework that analyzes dependencies between jailbreak attacks and defenses, proposing ensemble attack and defense approaches. \cite{prompt_taveekitworachai} explores the use of prompt evolution through examples to optimize prompts for specific tasks, demonstrating its applicability in toxicity classification. \cite{multi_sun} investigates multi-turn context jailbreak attacks, proposing CFA, a context-based method that dynamically integrates the target into contextual scenarios to conceal malicious intent. \cite{dpp_wan} introduces ASRA, a prompt optimization algorithm that utilizes determinantal point process (DPP) to select prompts based on both quality and similarity. \cite{jailbreaking_andriushchenko} demonstrates that even the most recent safety-aligned LLMs are vulnerable to simple adaptive jailbreaking attacks, highlighting the importance of adaptive strategies for red-teaming.

\section{Understanding the Underlying Mechanisms of Jailbreaking}

Understanding the underlying mechanisms of jailbreaking attacks is crucial for developing more effective defenses and robust LLM systems. 

\cite{towards_lin} investigates the behavior of harmful and harmless prompts in the LLM's representation space, hypothesizing that successful jailbreak attacks move the representation of harmful prompts toward the direction of harmless prompts. \cite{safedecoding_xu} introduces SafeDecoding, a safety-aware decoding strategy that amplifies the probabilities of safety disclaimers while attenuating those of potentially harmful content. \cite{jailbreaking_chao} proposes PAIR, an efficient black-box jailbreak algorithm inspired by social engineering attacks, demonstrating the effectiveness of iterative prompt refinement. \cite{human_das} explores converting nonsensical jailbreak prompts into human-understandable prompts by utilizing situational context, highlighting the need for defenses that consider context and meaning. \cite{pal_sitawarin} introduces PAL, a proxy-guided black-box attack that utilizes a surrogate model to optimize adversarial triggers. \cite{promptattack_shi} proposes PromptAttack, a method for constructing malicious prompt templates by investigating unfriendly template construction approaches. \cite{a_yang} explores the use of prompt-based adversarial attacks for both adversarial example generation and robustness enhancement. \cite{llm_yu} introduces LLM-Fuzzer, an automated fuzzing framework for discovering jailbreak vulnerabilities at scale. 

\subsection{Advanced Defense Strategies and Future Directions}

Several studies propose advanced defense strategies and explore new directions for securing LLMs against jailbreak attacks. 

\cite{cover_tan} presents COVER, a heuristic greedy attack algorithm that targets vulnerabilities in prompt-based learning. \cite{can_wu} investigates the potential of LLMs to automatically generate jailbreak prompts, highlighting the need for stronger defenses against automated attacks.  \cite{can_karkevandi} utilizes reinforcement learning to optimize adversarial triggers, enabling effective black-box attacks on aligned LLMs. \cite{jab_mehrabi} introduces JAB, a joint adversarial prompting and belief augmentation framework that aims to simultaneously probe and improve the robustness of LLMs.  \cite{rapid_shen} proposes RIPPLE, an optimization-based method inspired by psychological concepts to generate diverse and potent jailbreaking prompts. \cite{self_wang} introduces the concept of a semantic firewall and proposes self-deception attacks that can bypass such defenses by inducing LLMs to generate their own jailbreak prompts. \cite{fuzzllm_yao} presents FuzzLLM, an automated fuzzing framework that leverages templates and constraints to efficiently discover jailbreak vulnerabilities. \cite{gpt_ramesh} demonstrates that GPT-4 can jailbreak itself with high success rates using self-explanation, highlighting the importance of LLMs' reflective capabilities in security considerations. \cite{learning_qiang} investigates data poisoning attacks during instruction tuning, proposing a gradient-guided approach to learn adversarial triggers. \cite{guard_jin} introduces GUARD, a role-playing system that leverages existing jailbreak characteristics and a knowledge graph to generate new jailbreak prompts. \cite{virtual_zhou} proposes Virtual Context, a method that enhances jailbreak attacks by injecting special tokens to increase success rates. \cite{art_li} introduces ART, an automatic red-teaming framework for text-to-image models that leverages both vision and language models to identify vulnerabilities. 

\subsection{Prompt Optimization, Defense and Detection}

Prompt optimization plays a crucial role in both attack and defense strategies, leading to the development of specialized methods for enhancing the security of LLMs. 

\cite{rlprompt_deng} proposes RLPrompt, a reinforcement learning-based approach for optimizing discrete text prompts, demonstrating its applicability across various LLM types and tasks. \cite{asetf_wang} introduces ASETF, a framework that utilizes adversarial suffix embedding translation to generate coherent and understandable jailbreak prompts. \cite{efficient_niu} explores the use of visual modality to enhance the efficiency of LLM jailbreaking, demonstrating its effectiveness against various LLM models. \cite{enhancing_huang} proposes a reinforcement learning-driven query refinement framework to enhance LLM capability and robustness against jailbreak attacks. \cite{obscureprompt_huang} introduces ObscurePrompt, a method for generating stealthy jailbreak prompts by obscuring the original prompt using LLMs. \cite{baseline_jain} investigates the performance of baseline defense strategies against adversarial attacks, focusing on detection, input preprocessing, and adversarial training. \cite{mart_ge} proposes MART, a multi-round automatic red-teaming method that iteratively generates adversarial prompts and improves the safety of the target LLM. \cite{selfdefend_wang} introduces SelfDefend, a generic framework that employs a shadow LLM to protect the target LLM, showcasing its effectiveness against various jailbreak attacks. \cite{token_hu} proposes token-level adversarial prompt detection methods based on perplexity measures and contextual information. \cite{fight_mo} introduces Prompt Adversarial Tuning (PAT), a method that trains a prompt control prefix to defend against jailbreak attacks. \cite{robust_kim} proposes a robust safety classifier, Adversarial Prompt Shield, designed to detect and mitigate adversarial prompts, demonstrating its effectiveness in reducing attack success rates. 

\section{Benchmarking and Evaluation}

Benchmarking and evaluation are crucial for comparing different jailbreak attacks, defense strategies, and tracking progress in LLM security. 

\cite{jailbreakbench_chao} introduces JailbreakBench, an open robustness benchmark for jailbreaking LLMs, providing a standardized framework for evaluating attack and defense methods. \cite{not_greshake} investigates indirect prompt injection attacks on LLM-integrated applications, highlighting the vulnerabilities arising from the integration of LLMs into real-world systems. \cite{llm_helbling} proposes LLM Self Defense, a simple yet effective method that utilizes an LLM to screen generated responses for harmful content, demonstrating its efficacy against various attacks. \cite{defending_zhao} introduces Layer-specific Editing (LED), a defense mechanism that leverages the analysis of LLM layers to improve alignment against jailbreak attacks. \cite{ignore_schulhoff} launches a global-scale prompt hacking competition, collecting a large dataset of adversarial prompts and developing a taxonomy of attack types. 

\subsection{Addressing Specific Vulnerabilities}

Researchers are focusing on developing techniques to address specific vulnerabilities and refine the understanding of LLM security. 

\cite{evaluating_li} establishes a benchmark for evaluating the robustness of instruction-following LLMs against prompt injection attacks, highlighting the need to improve LLMs' understanding of prompt context. \cite{adversarial_liu} investigates adversarial attacks on ChatGPT, proposing prefix prompt mechanisms and external detection models for mitigation. \cite{fine_kumar} analyzes the impact of fine-tuning and quantization on LLM safety, demonstrating their potential to increase the success rates of jailbreak attacks. \cite{struq_chen} introduces StruQ, a defense mechanism that utilizes structured queries to separate prompts and data, improving resistance against prompt injection attacks. \cite{benchmarking_yi} proposes BIPIA, a benchmark for evaluating indirect prompt injection attacks, analyzing the underlying reasons for their success and proposing defense strategies. \cite{defending_hines} introduces spotlighting, a family of prompt engineering techniques to improve LLMs' ability to distinguish between multiple input sources, mitigating indirect prompt injection attacks. \cite{round_yung} proposes Round Trip Translation (RTT), a novel defense method that paraphrases adversarial prompts to enhance LLM's ability to detect harmful behaviors. \cite{ignore_ribeiro} introduces PromptInject, a framework for analyzing and generating adversarial prompts, highlighting potential vulnerabilities in deployed LLMs. \cite{codelmsec_hajipour} proposes CodeLMSec, a benchmark for evaluating the security of code language models, focusing on their susceptibility to generating vulnerable code. \cite{jailbreaking_liu} presents an empirical study on jailbreaking ChatGPT through prompt engineering, analyzing the distribution of jailbreak prompt patterns and their effectiveness in circumventing constraints. \cite{combating_chern} explores the use of multi-agent debate to mitigate adversarial attacks, demonstrating its potential to reduce model toxicity when LLMs engage in self-evaluation and feedback. \cite{reinforcement_wang} utilizes reinforcement learning to develop an LLM agent for automated attacks on other LLMs, enabling targeted attacks with precise control over the output.  \cite{tensor_toyer} introduces Tensor Trust, a dataset of human-generated adversarial examples for instruction-following LLMs, providing insights into real-world attack strategies. 

\section{Expanding the Scope of Red Teaming}

Recent studies aim to expand the scope of red teaming by incorporating ethical considerations, human factors, and broader societal implications. 

\cite{ai_greenblatt} focuses on improving the safety of AI systems despite intentional subversion, proposing robust protocols that utilize trusted models and human feedback for mitigating potential risks. \cite{jailbreaklens_feng} introduces JailbreakLens, a visual analytics system designed to facilitate the analysis and understanding of jailbreak attacks. \cite{llms_wu} presents a vision paper that emphasizes the potential of LLMs to defend themselves against jailbreak attacks by leveraging their own capabilities for detecting harmful prompts. \cite{student_llaca} explores the use of student-teacher prompting for red-teaming, aiming to improve the effectiveness of guardrails in LLMs. \cite{prompt_zhao} investigates the use of prompts as triggers for backdoor attacks, highlighting the vulnerabilities of prompt-based learning to malicious manipulations. \cite{cti4ai_nguyen} proposes CTI4AI, a system for generating and sharing AI security threat intelligence, emphasizing the need for collaborative efforts to enhance LLM security. \cite{formalizing_liu} presents a framework for formalizing and benchmarking prompt injection attacks and defenses, enabling a systematic evaluation of vulnerabilities and mitigation strategies. \cite{self_wang} proposes Self-Guard, a method that leverages the LLM itself to review and tag its responses for potential harmfulness, combining advantages of safety training and safeguards. \cite{jailbreaker_chen} introduces a moving target defense (MTD) approach to enhance LLM security, aiming to balance helpfulness and harmlessness by dynamically selecting outputs from multiple model candidates. 

\subsection{Exploring Biases and Societal Implications}

The ethical considerations of LLM development, including biases, misinformation, and societal impact, are receiving increasing attention. 

\cite{assessing_yang} assesses the adversarial robustness of LLMs, highlighting the influence of model size, structure, and fine-tuning strategies on their resistance to attacks. \cite{learn_xu} proposes a multi-agent attacker-disguiser game approach to mitigate refusal responses in LLM defense, enabling a more subtle defense mechanism. \cite{pku_ji} introduces PKU-SafeRLHF, a safety alignment preference dataset for Llama-family models, facilitating research on safety and harmlessness in LLMs. \cite{tiny_liu} proposes a prefix-model for mitigating toxic outputs by reconstructing prompts with minimal additional tokens, enhancing LLM safety and efficiency. \cite{pandora_deng} investigates RAG poisoning attacks on GPTs, showcasing the vulnerabilities of retrieval-augmented generation to adversarial manipulations. \cite{purple_zhou} presents PAD, a purple-teaming approach that combines red-teaming and blue-teaming techniques for iteratively improving LLM safety. \cite{towards_kruspe} explores methods for detecting unanticipated biases in LLMs, emphasizing the importance of Uncertainty Quantification and Explainable AI for transparency and fairness. \cite{a_mo} maps adversarial attacks against language agents, proposing a unified framework for understanding the security risks of increasingly complex LLM-based systems. \cite{latent_sheshadri} investigates latent adversarial training (LAT) to enhance the robustness of LLMs against harmful behaviors. \cite{pleak_hui} explores prompt leaking attacks, highlighting the risks of compromising intellectual property and enabling downstream attacks. \cite{prompt_agarwal} investigates prompt leakage effects in multi-turn LLM interactions, proposing defense strategies and evaluating their effectiveness. \cite{prsa_yang} introduces PRSA, a prompt stealing attack framework that analyzes input-output content to generate surrogate prompts, demonstrating its effectiveness against real-world prompt services. 

\subsection{Robustness, Human Factors and Ethical Considerations}

Researchers are exploring the impact of Unicode characters on LLM security and comprehension, developing personalized encryption frameworks for jailbreaking, and addressing exaggerated safety behaviors in LLMs. 

\cite{gptbias_zhao} proposes GPTBIAS, a comprehensive framework for evaluating bias in LLMs, leveraging the capabilities of advanced LLMs for assessment and interpretability.  \cite{she_chatrath} introduces a unique test suite of prompts for evaluating LLM alignment, focusing on fairness, safety, and robustness. \cite{arondight_liu} proposes Arondight, a red-teaming framework for VLMs that utilizes automated multi-modal jailbreak prompt generation, exposing vulnerabilities in toxic image generation and multimodal alignment. \cite{defending_cao} introduces RA-LLM, a robustly aligned LLM that incorporates a robust alignment checking function to defend against alignment-breaking attacks. \cite{autored_hasegawa} proposes AutoRed, a framework for automating red team assessments using reinforcement learning and strategic thinking, demonstrating its effectiveness in discovering network vulnerabilities.  \cite{the_zhang} examines the human factor in AI red teaming, highlighting the importance of considering human biases, blindspots, and potential psychological harms during red-teaming activities. \cite{are_abdelnabi} proposes activation-based methods for detecting task drift in LLMs, emphasizing the importance of analyzing internal model states for security. \cite{prompt_khomsky} investigates black-box attacks on defended LLMs, highlighting the challenges and significance of adversarial attacks in real-world scenarios. \cite{why_liang} analyzes prompt extraction threats, investigating the underlying mechanisms of prompt memorization and proposing defense strategies to mitigate such risks. \cite{whispers_evertz} explores confidentiality issues in LLM-integrated systems, proposing a secret key game for evaluating confidentiality and investigating robustness fine-tuning techniques for improving system resilience.

\section{Lessons Learned and Future Research}

Red-teaming research has significantly advanced our understanding of LLM vulnerabilities and the effectiveness of various attack and defense strategies. Several important lessons have emerged from this research, including:

\begin{enumerate}
    \item \textbf{LLMs are vulnerable to a wide range of attacks:} Even safety-aligned LLMs can be manipulated to produce harmful content, highlighting the need for continuous vigilance and robust defenses.
    \item \textbf{Black-box attacks pose significant challenges:} As many real-world LLM APIs are black-box, developing effective black-box attacks and defenses is crucial for securing LLM deployments.
    \item \textbf{Prompt engineering is crucial for both attack and defense:} The structure, content, and phrasing of prompts can significantly impact the success of jailbreak attacks and the effectiveness of defenses.
    \item \textbf{Multimodal and multilingual LLMs introduce new vulnerabilities:} The integration of visual and multilingual capabilities expands the attack surface and requires new defensive strategies.
\end{enumerate}

Future research directions in LLM red-teaming include:

\begin{itemize}

\item \textbf{Developing more sophisticated black-box attack and defense methods:}  As black-box LLMs are increasingly prevalent, research should focus on developing novel techniques that require minimal model access.
\item \textbf{Investigating the transferability and generalization of attacks:} Understanding the factors that influence the transferability of jailbreak attacks across different LLMs is crucial for developing robust defenses.
\item \textbf{Exploring the use of human-AI collaboration for red-teaming:} Combining human expertise with automated methods can lead to more effective and efficient red-teaming strategies.
\item \textbf{Addressing the ethical and societal implications of LLM red-teaming:} As red-teaming research becomes more sophisticated, it is important to consider the potential ethical and societal implications of developing and deploying powerful attack methods.
\end{itemize}

\section{Conclusion}

The growing field of LLM red-teaming has made significant progress in uncovering vulnerabilities and developing both attack and defense strategies. Continued research in this area is crucial for ensuring the responsible development and deployment of LLMs, fostering a safer and more secure AI ecosystem. By understanding the limitations and vulnerabilities of LLMs, researchers and developers can work collaboratively to develop robust safeguards and mitigation strategies that promote the beneficial and ethical use of these powerful language models.

These diverse applications of red-teaming principles highlight its importance not only for LLM security but also for a wider range of AI systems and real-world applications.

\bibliographystyle{plainnat}
\bibliography{redteam}

\end{document}